\def\year{2021}\relax
\documentclass[letterpaper]{article} 
\usepackage{aaai21}  
\usepackage{times}  
\usepackage{helvet} 
\usepackage{courier}  
\usepackage[hyphens]{url}  
\usepackage{graphicx} 
\usepackage{natbib}  
\usepackage{caption} 
\usepackage{amsmath}
\usepackage{amsthm}
\usepackage{amsfonts}
\usepackage{graphicx}
\usepackage{wrapfig}
\usepackage{tikz}
\usepackage{verbatim}
\usetikzlibrary{automata, positioning}

\newtheorem{definition}{Definition}

\newtheorem{proposition}{Proposition}
\newtheorem{corollary}{Corollary}

\title{Teaching to Learn:  Sequential Teaching of Agents with Inner States}
\author {
        Mustafa Mert \c Celikok \textsuperscript{\rm 1},
        Pierre-Alexandre Murena \textsuperscript{\rm 1}, 
        Samuel Kaski \textsuperscript{\rm 1,2} 
}
\affiliations {
    \textsuperscript{\rm 1} Helsinki Institute for Information Technology HIIT \\
Department of Computer Science, Aalto University, Helsinki, Finland \\
    \textsuperscript{\rm 2} Department of Computer Science, University of Manchester, United Kingdom \\
    mustafa.celikok@aalto.fi, pierre-alexandre.murena@aalto.fi, samuel.kaski@aalto.fi
}

\begin{document}

\maketitle

\begin{abstract}
In sequential machine teaching, a teacher’s objective is to provide the optimal sequence of inputs to sequential learners in order to guide them towards the best model. In this paper we extend this setting from current static one-data-set analyses to learners which change their learning algorithm or latent state to improve during learning, and to generalize to new datasets.
We introduce a multi-agent formulation in which learners’ inner state may change with the teaching interaction, which affects the learning performance in future tasks. In order to teach such learners, we propose an optimal control approach that takes the future performance of the learner after teaching into account. This provides tools for modelling learners having inner states, and machine teaching of meta-learning algorithms. Furthermore, we distinguish manipulative teaching, which can be done by effectively hiding data and also used for indoctrination, from more general education which aims to help the learner become better at generalization and learning in new datasets in the absence of a teacher.
\end{abstract}

\section{Introduction}

Pedagogical systems are intelligent systems in which an agent, called the \textit{teacher}, transmits data to a second agent, called the \textit{learner} in order to help them learn a target concept~\cite{shafto2014rational}. Intelligent systems which aim to help human users build statistical models of their data can be seen as pedagogical systems, in which the user takes over the role of the learner. The main problem for the teacher in this case would be to optimize its sequential interaction with the learner in order to help them build a better model of their data.

\emph{Machine teaching} addresses the problem of finding the best training data that can guide a learner, human or machine alike~\cite{patil2014optimal,chen2018understanding}, to a target model with minimal effort \cite{zhu2015machine,GOLDMAN199520}. 
However, conventional machine teaching considers a restricted class of learners which have fixed inductive biases (e.g. parameter initialization, model family, network architecture, variable selection etc.) and hyper-parameters. 
For sequential teaching interactions, 
this assumption
means that the learner cannot update their inductive biases during the learning process, which human learners and many machine learning methods (such as meta-learning) can actually do based on the data they have seen to achieve better generalization amongst similar learning tasks. 

\begin{figure}
\center 
\includegraphics[scale=0.5]{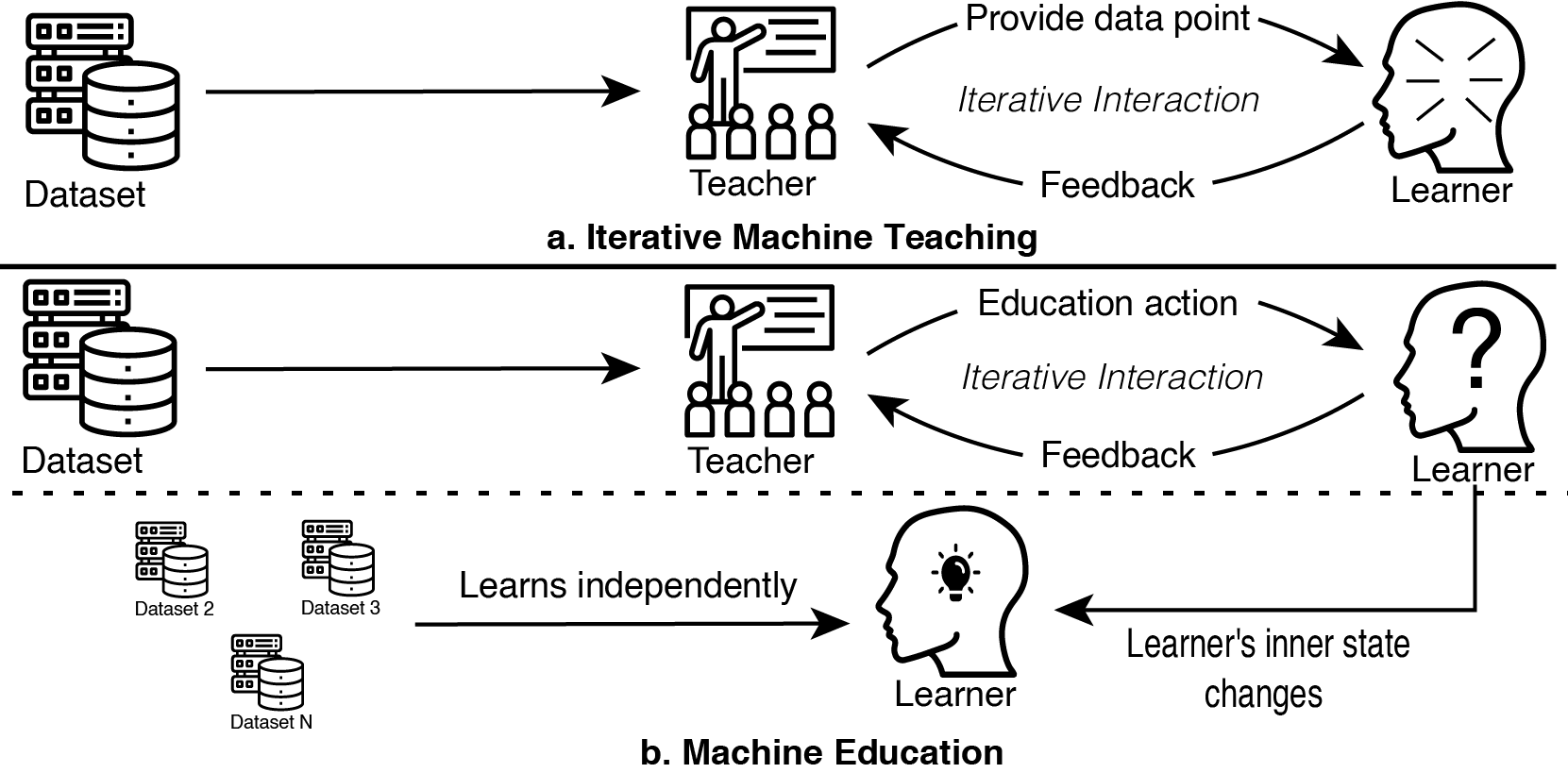}
 \caption{a) Iterative machine teaching: Teacher provides data points iteratively from a single dataset to a fixed learning algorithm. b) Machine education: More generally, the learner has inner states that can tune its learning algorithm. The teacher takes education actions which changes the learner's inner state and makes them better at learning independently with new datasets. }
 \label{fig:illustrative}
\end{figure}

If the learner's initial biases are unsuitable for the task and cannot change, we show that the teacher may then need to hide some data-points from the learner in order to make them learn a better model than the one that would be inferred from the whole dataset. 
This teaching strategy is close to data-poisoning~\cite{mei2015using} and may be seen as an undesirable and patronising behaviour which attempts to manipulate the learner. 
However, considering that the learner’s biases can change and be influenced by the teacher induces a completely different teaching strategy: helping the learner refine their \emph{inner state}, essentially teaching them better biases and hyper-parameters, before they can learn the model. 
This empowers the learners by teaching them to perform better 
during the learning phase, with assistance of the teacher, but also in future tasks, even in the absence of the teacher.
We refer to this deeper goal as \emph{machine education}  (Figure~\ref{fig:illustrative}); it can be done by allowing the teacher to use a potentially wider set of actions than just choosing data points.

In this paper, we formalize the problem of education as a two-player game involving two agents, a \emph{learner} and a \emph{teacher}, and we consider the problem of finding the optimal teaching strategy for the teacher. 
To this end, we model the learner as having a latent inner state, which represents their understanding of the modelling process in the form of their biases and hyper-parameters. 
This latent state changes over time, as a result of the teacher's actions, in addition to the rest of the environment.
The task of the teacher consists not only in guiding the learner toward selecting a model close enough to the best model of the data possible, through a sequence of interactions, but also in guaranteeing that the learner will be able to select good models without supervision in future similar tasks. 
To do so, the teacher needs to lead the learner to an inner state which guarantees a good understanding of the modeling process.
Our main contributions are: (i) We generalize sequential machine teaching to a setting where the learner has an inner state which affects their preferences over models, and evolves over time in response to the teacher's actions. (ii) We show that, when the learner's inner state is static and sub-optimal, optimal teaching is possible only at the price of some manipulation, defined in detail below. 
(iii) We show that augmenting machine teaching by considering the teacher's influence on the learner's inner states allows the teacher to avoid manipulative strategies and help the learner learn to perform better later, even in the absence of the teacher.
Before introducing these results, we will first present an example of teaching to humans which gives the intuitions behind our framework. This example will then be formalized and used for empirical validation. For completeness, we also propose some results for another application, of teaching to machines. 


\section{An Illustrative Example}
\label{sec:motivating-example}

Consider an intelligent system that is designed to help its users build linear models, such as $\mathbf{Y} = \mathbf{X}\mathbf{\xi} + \epsilon$ with $\epsilon \sim \mathcal{N}(\mu, \sigma^2)$, for their data $\mathcal{D} = \lbrace (X_i, Y_i) \rbrace_{i=1 \dotsc n}$, where $X_i \in \mathbb{R}^d$ and $Y_i \in \mathbb{R}$. Building linear models of data is a ubiquitous task across all science disciplines. An important aspect of the linear models is their interpretability: the coefficients $\xi$ are easy to interpret in terms of describing linear relationships between inputs $\mathbf{X}$ and outputs $\mathbf{Y}$. 

In this scenario, the intelligent system can model its user as a learning algorithm which, given data as input, produces a linear model as output. The task of the intelligent system is to help the learning algorithm converge to a good model with minimal effort. Evidently this setting can be modelled as machine teaching where the system is the teacher, and the user is the learner.

However, often the users of such systems do not have advanced knowledge of statistical model building. For instance, consider the task of selecting which covariates from $\lbrace 1, \dotsc, d \rbrace$ to include in the linear model. The intelligent system (teacher henceforth) helps the user (learner henceforth) by suggesting which covariates to include. The learner can then accept or reject the suggestions throughout the sequential interaction. A learner who does not know about the effects of collinearity, for instance, on model interpretability and uncertainty, may choose to include strongly collinear covariates into the model if they are correlated to the output. If we apply conventional machine teaching to this case by treating covariate suggestions as data, the teacher's optimal behaviour would be to avoid suggesting collinear covariates. Such a strategy is intuitively optimal in terms of the model finally obtained, since it prevents the aforementioned learner from including collinear covariates. 
However, it is important to recognise that the model would then be built by effectively hiding from the learner information that they could misinterpret:
Had the learner observed the entire dataset by themself, they would have included collinear covariates and ended up with a different model. This is not satisfying since it means the learner will not be able to choose a good model for future datasets, unless the teacher is there to supervise them.
We argue that this discrepancy between the model built when following the supervision of the teacher, and the model built without supervision when given the whole dataset, can be interpreted as resulting from a manipulative teaching strategy. 

This can be avoided by allowing the teacher to influence the modelling biases and preferences of the learner, corresponding to their inner state. A teacher able to infer whether the learner's modelling preferences disfavour collinearity, and equipped with tutoring actions (which can communicate to the learner the negative effects of collinear covariates), could consider an educative strategy instead: help the learner understand the notion of collinearity, and therefore change their inner state for the better. This means that, in the future, the learner will be able to prefer minimal collinearity amongst covariates when building linear models. 


The framework of machine education, introduced in the upcoming section, formalizes the intuitions presented here. In particular, we will demonstrate in Proposition~\ref{prop:impossibility-teaching} below that unless the teaching aims at changing the learner’s inner state, the teacher’s choices are either to manipulate (in a sense that will be defined in next section) or end up with sub-optimal learning results. 
A crucial insight of our work is that, by taking the future modelling performance of the learner in the teacher's absence into account, the teacher can plan the education to lead to beneficial changes in the learner's inner state, utilizing whatever actions are available.
The optimality of such a teaching policy will be exposed in Proposition~\ref{prop:optimality-education}.




\section{The Model of Machine Education}
\label{sec:machine-education}

In this section, we formalize the intuitions discussed above and introduce the general setting of machine education.

\subsection{Sequential Teaching of Models}
\label{subsec:sequential-teaching}

\paragraph{Modeling task.} We formalize the overall problem as a modelling task, in which a model $\theta$ has to be learnt to describe a dataset~$\mathcal{D}$. We denote by $\Theta$ the class of models for this task and endow $\Theta$ with a discrepancy function $d(.,.): \Theta \times \Theta \mapsto [0, \infty)$. 
We assume that the discrepancy function is such that $d(\theta_1, \theta_2) = 0$ if and only if $\theta_1 = \theta_2$. 
In the case of probabilistic modelling, $\theta$ is the posterior distribution over model parameters and $d$ is a discrepancy measure between probability distributions (e.g. KL divergence). 


\paragraph{Multi-agent model.} We consider two agents: a learner
and a
teacher. The teacher has better inductive biases than the learner, and therefore can identify a better model $\theta^* \in \Theta$ than the learner. The learner aims to select a model to describe the data.
The interaction between the two agents is modelled as a sequential leader-follower game. At each time step $t$, the teacher selects an action $a_t \in \mathcal{A}$ to perform, the learner responds with an action $b_t \in \mathcal{B}$ and updates its selected model $\theta_t$. In essence, every action of the teacher $a_t$ can be seen as suggesting a model or a hypothesis to the learner. The learner may accept or reject this suggestion, or simply ignore it when updating its model. 

In this paper, we take the position of the teacher and aim to find the optimal sequence of actions minimizing the distance $d(\theta_T,\theta^*)$ for a certain horizon~$T$.


\paragraph{Learner's type space.} 
In game theory an agent's type is a representation of its beliefs and objectives. A type space is the set of all agent types considered in a game. 
In our context, the learners' type space $\mathcal{Z}$ can be represented as the product of a function space $\mathcal{F}$ and a set of algorithms $\Pi$ (as in machine teaching, an algorithm is a function mapping a dataset to a model).
A \emph{learner's inner state} is then an element of this product space, defined as the tuple $z = (f, Alg(D; f, \Theta))$, where $f$ is a real-valued function inducing a preference ordering in $\Theta$ for the data $D$ based on the learner's biases. In the case of probabilistic learners, $f$ can be chosen as the prior density over models, and $Alg(D; f, \Theta)$ denotes an algorithm (e.g. Bayesian learning rule, variational inference, gradient descent) the learner uses to build a model of the dataset $D$, parameterised by the model space $\Theta$ and modelling preferences. The probability of a learner with inner state $z$ responding to $a_t$ with $b_t$ is denoted as $\pi_\ell(b_t | a_t, \theta, z)$. 
We refer to the supplementary material for an in-depth discussion about $\mathcal{F}$ and $\Pi$. 
Introducing the history $H_t = (a_1, b_1, \dotsc, a_t, b_t)$, we assume that a learner's inner state at a given time $z_t$ evolves according to the transition probabilities $p(z_{t+1} | z_t, H_t)$ which will be called the \emph{inner state dynamics}. We assume the teacher knows the parametric form of the transition probabilities; this assumption can be easily relaxed.



\paragraph{Model of the teacher. } Given a type space for learners, the teacher's decision-making can be modelled as a POMDP $\mathcal{M} = (\mathcal{S}, \mathcal{A}, \mathcal{T}, \mathcal{R}, \Omega, \mathcal{O})$, where $\mathcal{S} = \Theta \times \mathcal{Z}$ is the state space, $\mathcal{A}$ the space of actions introduced earlier,  $\mathcal{T}$ the transition kernel, $\Omega$ the set of observations, $\mathcal{O}$ a set of conditional observation probabilities and $\mathcal{R}$ a cost function\footnote{In our applications, we will use control-theoretic cost minimization instead of reward maximization. These two formulations are equivalent.} that will be discussed in conclusion of Section~\ref{subsec:non-manipulative}. A state $s \in \mathcal{S}$ is composed of two components $s = (\theta, z)$, where $\theta$ is the model selected by the learner and $z$ is the learner's inner state. The $z$ cannot be directly observed but can be inferred from the learner's policy $\pi_\ell(b_t | a_t, s)$, therefore $\Omega = \mathcal{B}$ and $\mathcal{O}=\pi_\ell$. 

\subsection{Cost for a Non-Manipulative Teacher}
\label{subsec:non-manipulative}


Machine education consists, for the teacher, in helping the learner select the best possible model $\theta^* \in \Theta$ to describe the dataset $\mathcal{D}$. 
We now identify a desirable property for the teacher, which is to avoid manipulating the learner. 
We formalize the notion of \textit{manipulation} as follows: Manipulation level measures the discrepancy between the model $\theta$ learned by a learner during an education process and the model that the learner would infer from the whole dataset, without assistance from a teacher. 

\begin{definition}[Manipulation and Enlightened inner state] The \textbf{manipulation level} on data $\mathcal{D}$ of a learner of type $z \in \mathcal{Z}$ toward model $\theta \in \Theta$ is defined as $Manip(z, \mathcal{D}, \theta) = d(Alg_z(\mathcal{D}; f_z, \Theta), \theta)$. Additionally, we say that an inner state $z \in \mathcal{Z}$ is \textbf{enlightened} for dataset $\mathcal{D}$ toward model $\theta$ if $Alg_z(\mathcal{D}; f_z, \Theta) = \theta$ (or equivalently $Manip(z, \mathcal{D}, \theta) = 0$).
\end{definition}

In the following propositions (proofs in supplementary material), we demonstrate the importance of considering the possibility of inner state transitions to provide optimal and non-manipulative teaching. We always consider the dataset $\mathcal{D}$ fixed and denote by $\mathcal{Z}^*(\theta) \subset \mathcal{Z}$ the set of all enlightened inner states for data $\mathcal{D}$ towards the model $\theta$.

\begin{proposition}
Suppose that the initial inner state of the learner $z_0$ is not enlightened ($z_0 \not\in \mathcal{Z}^*(\theta^*)$) and that, for all $n > 0$, $p(z_n \in \mathcal{Z^*} | z_0) = 0$. 
Then for any $n > 0$, with probability 1 at least one of the two following statements is true: (1)  $Manip(z_n, D, \theta^*) > 0$ or (2) There exists a model $\theta^\prime$ such that $d(\theta^\prime, \theta^*) < d(\theta_n, \theta^*)$ and $p(\theta_n = \theta^\prime | z_0) > 0$.
\label{prop:impossibility-teaching}
\end{proposition}

Proposition~\ref{prop:impossibility-teaching} shows that a teacher who would not enlighten the learner (for instance by not triggering any change in learner's inner state) is necessarily limited to either being manipulative or being sub-optimal. 
This impossibility result applies in particular to machine teaching techniques which allow the teachers to alter the data distribution by filtering out samples or providing data that is inconsistent with the data distribution as shown by~\citet{peltola2019machine}. 

The following proposition states that, when inner states can be influenced by the teacher, the teacher can guide the learner towards an inner state where $\theta^*$ could be retrieved without assistance, essentially allowing the teacher to avoid manipulating the learner.

\begin{proposition}
Suppose that the learner's inner states $(z_t)$ are observed by the teacher. If there exists an enlightened inner state $z^*$ such that $p^* = p(z_t = z^* | z_0) > 0$ for $t$ large enough, then there exists a policy $\pi$ for the teacher such that, with probability $p^*$, $\theta_T$ obtained by teaching interaction is optimal ($\theta_T = \theta^*$) and non-manipulative ($Manip(z_T, D, \theta) = 0$) for some $T > 0$.
\label{prop:optimality-education}
\end{proposition}

Even though Proposition~\ref{prop:optimality-education} assumes that the learner's inner state is observed, it is sufficient in practical applications that the inner state can be inferred based on the interaction data.

These two propositions imply that optimal teaching can be made non-manipulative by allowing the teacher to help the learner switch from one inner state to the other. 
Here, non-manipulative teaching means that the learner is eventually able to make the same choice of a model without any supervision.

Another desirable property of learning would be the ability for the learner to perform correctly on new datasets. 

\begin{corollary}
Under the conditions of Proposition~\ref{prop:optimality-education}, let $(\mathcal{D}^\prime, \theta^{\prime*})$ be a dataset and an associated model, and suppose that $z^* \in \mathcal{Z}^*(\theta^{\prime*})$. Then $Alg_{z_T}(\mathcal{D}^\prime; f_{z_T}, \Theta) = \theta_T$.
\end{corollary}

This observation shows that allowing tutoring actions in teaching does not only guarantee optimality of the modeling for the task of interest, but also for any similar task, where similarity is defined by a common enlightened inner state. We notice here a strong connection with meta-learning. Indeed, meta-learning, also commonly referred to as \textit{learning to learn}~\cite{thrun2012learning, vanschoren2019meta}, is a learning paradigm in which the meta-learner aims to help a learner configure a proper algorithm to solve various similar tasks. In our context, the evolution of $z$ during the interactions with the teacher can be interpreted as a meta-learning algorithm learning the meta-parameters.

\paragraph{Choice of the teacher's cost for non-manipulative teaching.} 
We model the teacher's decision-making as a multi-objective POMDP with three possible objectives: (O1) Assist the learner to select the optimal model $\theta^*$ for $\mathcal{D}$; (O2) Make the learner able to select the best model $\theta^* \in \Theta$ for $\mathcal{D}$ without assistance; (O3) Make the learner able to select the optimal model for tasks similar to $\mathcal{D}$ without assistance. Objective (O1) can be achieved without considering (O2), but could be manipulative (Proposition~\ref{prop:impossibility-teaching}). Also, objective (O2) implies objective (O3) if we can guarantee that the tasks share a common enlightened inner state. 
The corresponding costs are given by: (O1) the final model discrepancy $d(\theta_T, \theta^*)$; (O2) the final manipulation level $Manip(z, \mathcal{D}, \theta)$; and (O3) model discrepancy for related tasks $\mathcal{D}^\prime$: $\sum_{\mathcal{D}^\prime} d(Alg_{z_T}(\mathcal{D}^\prime; f_{z_t}, \Theta), \theta^*(\mathcal{D}^\prime))$. Equivalently, we can consider that $\mathcal{D}$ is a future task and objective (O2) is already included in objective (O3). 
We map the three-objective cost to a single objective function $g_T(z_T, \theta_T)$ with a \emph{linear scalarization function} with a parameter $u=(u_1, u_2)$ (controlling which objective the teacher should prioritize more). 
\begin{equation}
    g_T = u_1 d(\theta_T, \theta^*) + u_2 \sum_{\mathcal{D}^\prime} d(Alg_{z_T}(\mathcal{D}^\prime; f_{z_t}, \Theta), \theta^*(\mathcal{D}^\prime))
    \label{eqn:terminal-cost}
\end{equation}



\section{First Application: Interactive Variable Selection with Users}
\label{sec:application-mme}
We now apply our framework to the setup presented in Section~\ref{sec:motivating-example} where the teacher helps a (simulated) user build linear models.

\paragraph{Description of the task.} 
The goal of the learner is to choose which variables to include in the linear model. A variable can be excluded from the regression by setting its weight to zero as $\xi^i = 0$. Thus, the model space is the space of d-dimensional binary vectors $ \Theta = \{0,
1\}^d$ with each dimension denoted as $\theta^i = \mathbb{I}(\xi^i \neq 0)$ where $\mathbb{I}$ is the indicator function. 
At each time-step the teacher can select a variable $i \in \lbrace 1, \dotsc, d \rbrace$ from the dataset to display or provide explicit explanations about the design of linear models (which corresponds to an action called $tutor$). Therefore, the action space of the teacher is $\mathcal{A} = \lbrace 1, \dotsc, d \rbrace \cup \lbrace tutor \rbrace$. 
At time $t$, the learner observes action $a_t$ from the teacher and picks a response $b_t \in \lbrace 0, 1 \rbrace$ 
corresponding to rejecting or accepting the suggestion of the teacher. In case $a_t = i \in \lbrace 1, \dotsc, d \rbrace$ is not a tutoring action, the learner updates the model $\theta_t$ based on whether they accepted to include the suggested variable or not, therefore $\theta_t^i = b_t$.

\paragraph{Learners' type space.} 
When making modelling decisions, different learners pay attention to different statistics in the data and the model, but to extents unknown to the teacher. Based on this observation, the teacher formulates the learner's modelling preferences as functions of the form $f(\phi(\theta, a);\mathbf{w}_z) = \mathbf{w}_z^T \phi(\theta, a)$ where $\phi(\theta, a)$ is an embedding of the statistics, for a model suggested by the teacher through action $a \in \lbrace 1, \dotsc, d \rbrace$. $\mathbf{w}_z$ is an unknown weight vector capturing how much the learner pays attention to each of them. Therefore the space of preference functions $\mathcal{F}$ (introduced in Section~\ref{subsec:sequential-teaching}) is defined as set of linear functions from the embedding space to $\mathbb{R}$. Since the learner is doing linear regression, the space of algorithms $\Pi$ consists of a single algorithm which performs the regression.

The feature map $\phi$ (embedding) encodes the quantities of interest to the learner, i.e. here the correlation of the shown variable to the output, and (maximal) correlation with already included variables as 
$\phi(a_t, \theta_{t-1}) = (|corr(a_t,Y)|, \max_{j : \theta_{t-1}^j \neq 0} |corr(a_t, j)|)$.

With this type space, the general policy of the learner is then given by:
\begin{equation}
    b_t | a_t, z_t \sim Bernoulli\left(\sigma(f_{z_t}(\phi(a_t, \theta_{t-1})))\right)
    \label{eqn:mme-learner}
\end{equation}
where $z_t \in \mathcal{Z}$ denotes the learner's inner state at time $t$ and $f_{z_t}$ corresponds to its preference function $f$.

As discussed in Section~\ref{sec:motivating-example}, two behaviors can be observed depending on whether the learner knows collinearity. Formally, we observe that this corresponds to the decomposition of $\mathcal{Z}$ into two subspaces:
$\mathcal{Z} = \mathcal{Z}^{(0)} \cup \mathcal{Z}^{(1)}$.  
The subspace $\mathcal{Z}^{(0)}$, associated to $\mathcal{F}^{(0)} = \lbrace f: x \mapsto \mathbf{w}^T x : \mathbf{w} = (w_1, 0), w_1 \in \mathbb{R} \rbrace$, describes naive learners who do not understand collinearity, whereas $\mathcal{Z}^{(1)}$, associated to $\mathcal{F}^{(1)} = \lbrace f: x \mapsto \mathbf{w}^T x : \mathbf{w} = (w_1, w_2), w_1 \in \mathbb{R}, w_2 < 0 \rbrace$, describe enlightened learners who understand collinearity and would avoid it.

\paragraph{Learner's inner state dynamics.} 
Based on our simplifying assumptions, only the action $a_t = tutor$ can cause changes in the inner state, with probability $\eta$, resulting in the following dynamics:
$p(z_{t+1} \in  \mathcal{Z}^{(1)} | z_t \in  \mathcal{Z}^{(0)}, a_t \neq tutor) = 0$, $p(z_{t+1} \in  \mathcal{Z}^{(1)} | z_t \in  \mathcal{Z}^{(0)}, a_t = tutor) = \eta$ and $p(z_{t+1} \in  \mathcal{Z}^{(1)} | z_t \in  \mathcal{Z}^{(1)}) = 1$. 
As a consequence,
the data-generating process for 
feedback $b_t$ is a Markov-switching model \cite{hamilton1989new}.



\paragraph{Teacher's cost.} To complete the definition of the POMDP for the teacher, we define a stage cost function $g: \mathcal{A} \rightarrow [0, \infty)$. In this application, we take $g(a)$ constant for all $a \in \lbrace 1, \dotsc, d \rbrace$, but it would be possible to generalize to variable-specific costs, implying that some features are more difficult to assess by the learner. Also, we assume that the cost of the tutoring action $g(tutor)$ is higher than the cost of a variable recommendation. We complete the teaching with the terminal cost introduced in Equation~\ref{eqn:terminal-cost}.


\begin{figure}

  \begin{center}
    \includegraphics[width=0.4\textwidth]{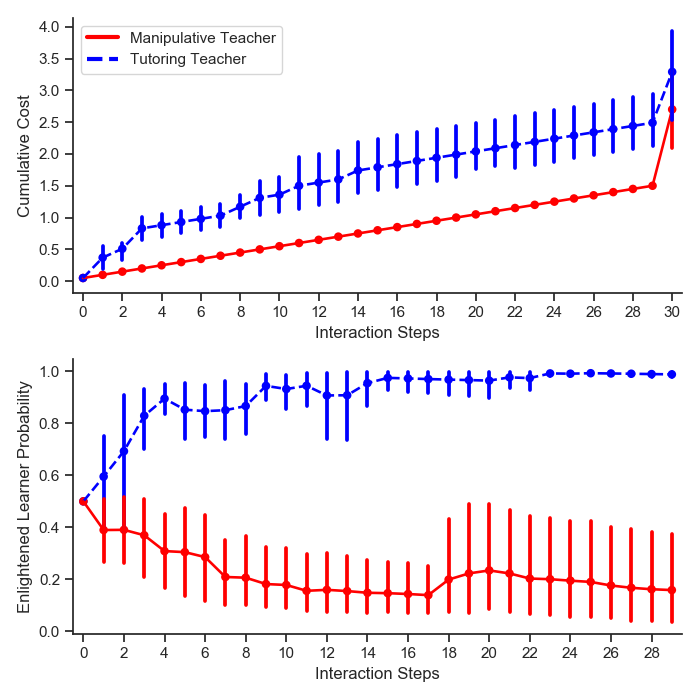}
  \end{center}
  \caption{Comparison of a manipulative and a tutoring teacher (bars indicate 95\% CI). Top: when only the performance on the current dataset matters for the terminal cost observed at the last time-step, the manipulative teaching (red) policy is cost-optimal and there is no need to tutor. Bottom: tutoring teacher (blue) leads to type changes from naive to enlightened.}
  \label{figure_1_experiment_1}
\end{figure}


\paragraph{Algorithm.} In the POMDP with state $s = (\theta, z)$, the model $\theta$ is observed, but the learner's type $z$ is not. 
It can be inferred from the posterior $p(z_t | H_t)$, the detailed expression of which is provided in the supplement.
We solve this POMDP by using problem approximation \cite{bertsekas2019reinforcement} and turning this into a simpler fully-observed stochastic dynamic programming problem by repeating the following process: We take posterior expectations $\Bar{\alpha}_t, \Bar{\mathbf{w}} | H_t$ and sample the space $\mathcal{Z}^{(n_t)}$ of the learner types by $\Tilde{n}_t \sim Bernoulli(\Bar{\alpha}_k)$. We then use rollout by simulating the decision trajectory with a fixed parameter $\Bar{w}$, based on the learner's state transition dynamics and policy given by Equations~\ref{eqn:mme-learner}. The optimal solution for this problem is selected as action $a_{t+1}$. After getting learner's feedback $b_{t+1}$, the belief $p(\alpha_{t+1}, w | H_{t+1})$ is updated and the process is repeated.

\subsection{Experimental Results}

\paragraph{Setup.} We use the data generation method provided by \citet{ghosh2015bayesian} for comparing method performances in collinear datasets, and generate random regression datasets with 10 independent and 15 collinear variables (details in the supplementary materials). Such high degree of collinearity is a typical feature of large-panel macroeconomic data \cite{de2008forecasting}. All results have been replicated with 10 random seeds and we present averaged values with 95\% confidence intervals (CI). 
We simulate the learner's behaviour using the presented model (policy~\ref{eqn:mme-learner} and learner's inner state dynamics).
Unless stated otherwise, the value for $\eta$ is $0.5$. Sensitivity analysis is in the supplement. The optimal variable selection strategy is to include all independent variables, and choose only one from the collinear variables. Once the variable selection is done, the learner pays a unit cost ($1.0$) for each missed independent variable and every extra collinear variable selected, which corresponds to a penalty $d(\theta, \theta^*)$, $d(.,.)$ being the Hamming distance


\paragraph{Experiment 1: Manipulative teaching is optimal for the current dataset. } In typical iterative teaching, the goal is to guide the learner into the best possible model with minimal cost for a given dataset, which corresponds to the scalarization $u_1=1, u_2=0$ (only the current dataset is considered in the terminal cost). The cumulative cost in Figure~\ref{figure_1_experiment_1} shows the performance of our rollout method (blue) against a teacher who never chooses to tutor when the scalarization is given by $u_1=1, u_2=0$ (red). 
According to Proposition~\ref{prop:impossibility-teaching}, such a teacher is expected to be manipulative. Due to the rollout approximation, our method chooses to educate in multiple time-steps and thus has a higher cumulative cost. Evidently, the optimal policy in this setting should never tutor, and can simply manipulate the learner by never showing a second variable from the collinear group.

\begin{figure}
  \begin{center}
    \includegraphics[width=0.4\textwidth]{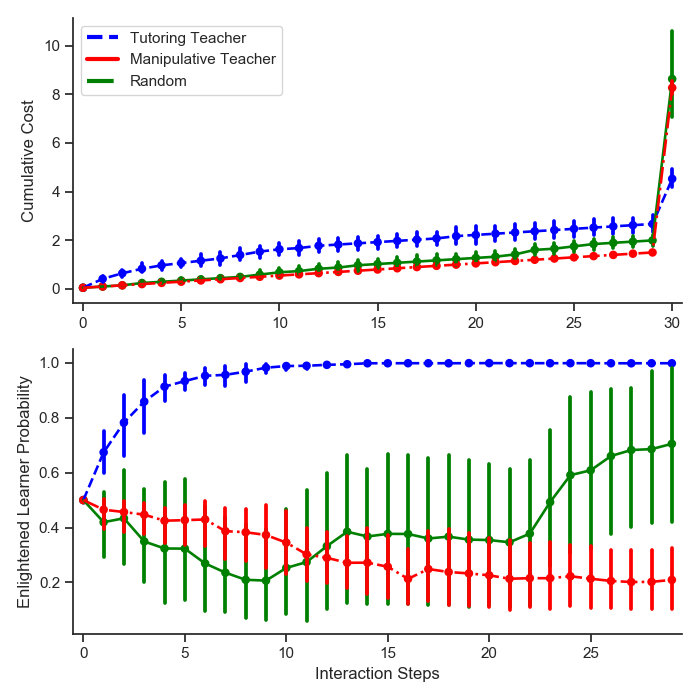}
  \end{center}
  \caption{Comparison of mean teaching performances for manipulative, tutoring and random teachers with 95\% CI.
  Top: machine education induces a lower cumulative cost than manipulative teaching, since an estimate of the learner's independent learning performance after interaction is included in the terminal cost observed at the last time-step.
  Bottom: machine education leads to a type change early on, whereas  manipulative teaching does not cause any type changes. }
  \label{fig:claim_4}
\end{figure}
\paragraph{Experiment 2: Manipulative teaching leads to low performance in independent learning.} In order to evaluate how the two types of learners perform without the presence of a teacher, we generated 10 test datasets, having the same degree of collinearity as the sets used for teaching in Experiment 1. 
We observe that, on 10 datasets sampled from the task distribution, enlightened learner gets a mean terminal cost of 2.18 (stdev 0.44), while naive learner gets 12.34 (stdev 0.29). 
As expected, in the absence of a teacher, the enlightened learner performs much better than the naive one since it takes collinearity into account. 


\paragraph{Experiment 3: Including an estimate of the independent learning performance to the cost leads to enlightenment.} 
We generated 10 additional datasets from the same generation process with the same degree of collinearity. Differently from test datasets, we use these to estimate the mean of future regret, the second term 
$g_2$ in the teacher's cost formulation.
This term serves as an estimator of the learner's independent performance on similar datasets, when the teacher is not present. We set $u_1=0.5, u_2=0.5$, hence the current and future performances are considered equally important. As seen in Figure~\ref{fig:claim_4}, this makes the tutoring teacher the best choice compared to the manipulative and random teachers: the cumulative cost of the tutoring teacher outperforms all, and the learner is tutored to switch to enlightened, as seen in our model's confident inference of the learner type. Since the learner becomes enlightened, its generalization performance improves drastically as shown with Experiment 2. Details on how the tutoring teacher method induces inner state changes and how our model detects these changes in an episode are provided in the supplementary materials for two different values of $\eta$.


\section{Second Application: Teaching Online Meta-Learners}

We next apply our framework to teaching an online meta-learner to learn a good initialization.

\paragraph{Description of the task.}

Consider a learning task $\mathcal{T} \sim P(\mathcal{T})$ represented by a tuple $\mathcal{T}= (\mathcal{D}^{tr}, \mathcal{D}^{test})$ consisting of a training and a test dataset. All learning tasks that come from $P(\mathcal{T})$ have some common statistical properties. If a learner can exploit these common properties via inductive biases, it can generalize to new tasks faster. The goal of meta-learning is to learn these inductive biases from a set of tasks.

Model-agnostic meta-learning (MAML)~\cite{finn2017model} is a general framework for meta-learning applicable to any model that is trained by gradient descent. The goal of MAML for neural networks (NN) is to learn an initialization of the NN parameters $\theta_0$ that quickly leads to good models for any task from $P(\mathcal{T})$. Initial model $\theta_0$ can be seen as some form of modelling preferences and biases since the starting point on the parameter space indirectly induces a preference over the model space $\Theta$ 
due to finite data.

In order to learn a good $\theta_0$, MAML uses a set of task samples $\{\mathcal{T}_i\}_{i=1,...,M}$ and minimizes the meta-learning loss $F(\theta) = \frac{1}{M} \sum_{i=1}^M {\mathcal{L}(Alg(\mathcal{D}^{tr}_i, \theta),\mathcal{D}^{test}_i)}$, where $\theta \in \Theta$ corresponds to the parameters of the model. An online variant of this problem has been studied in~\cite{finn2019online} where the meta-learner can get tasks only one-by-one.

In this section, we consider the new problem of teaching online meta-learners a good initialization $\theta^*_0$.\footnote{We implicitly assume here that the teacher cannot transmit the value $\theta^*_0$ to the learner, for instance in case the model would be too costly to transmit.}

\paragraph{Learner's type space and inner state dynamics.} For the type space of online meta-learners, the space of algorithms $\Pi$ (see Section~\ref{subsec:sequential-teaching}) consists of a single learning algorithm $Alg$ which is stochastic gradient descent. In this setting, the space of modelling preference functions $\mathcal{F}$ is implicit, yet we can assume $\mathcal{F}$ is parameterized by $\theta_0$ since each initialization induces a preference. Thus instead of $\mathcal{F}$ we will use $\Theta$. The meta-learner has no choice but accepting the dataset ($\pi_\ell(b_t=1) = 1$) and updates $\theta_0$ by using the sublinear regret method introduced in~\cite{finn2019online} and called \textit{follow the meta-leader}:
$
    \textbf{FTML}(\theta_t, \{\mathcal{T}_i\}_{i=1,...,t})= \text{arg} \min_{\theta}\left\lbrace\frac{1}{t}\sum_{k=1}^t{\mathcal{L}(Alg(\mathcal{D}^{tr}_k, \theta),\mathcal{D}^{test}_k)} \right\rbrace
$

\paragraph{Model of the teacher.} The tutoring actions of the teacher correspond to the choice of a task to present to the learner: $\mathcal{A} = \lbrace \mathcal{T}_i \rbrace_{i=1,...,M}$. since they directly affect $\theta_0$.
Once a task $\mathcal{T}$ is chosen, the entire training dataset $\mathcal{D}^{tr}$ for $\mathcal{T}$ is used. Then the teacher has only the Objective (O3) to consider. We choose to model the cost as the Euclidean distance to $\theta^*_0$ denoted by $d(\theta, \theta^*_0)$.



\paragraph{Algorithm.} 
The education interaction again defines a sequential leader-follower game. The teacher, as the leader, chooses which task to add to the current sequence of tasks. The learner responds by applying the FTML algorithm to update its initialization $\theta_0$. The Stackelberg equilibrium for the stage game at time $t+1$ can be computed by solving the following bi-level optimization task:
\begin{align*}
\min_{\mathcal{T}} \  d(\theta, \theta^*_0) \quad
\textrm{s.t.} \quad  \theta \in \textbf{FTML}(\theta_t, \{\mathcal{T}_i\}_{i=1,...,t} \cup \mathcal{T})
\end{align*}

FTML is a myopic follower and the dynamics are fully controlled by the leader's policy. Either of these properties sufficiently admits a dynamic programming solution to the computation of a strong Stackelberg equilibrium \cite{bucarey:hal-02144095}. Our rollout approximation uses one-step look-ahead minimization and chooses the task that minimises $d(\theta_{t+1}, \theta^*)$ at time $t$ by applying the difficulty and usefulness decomposition given by \citet{liu2017iterative} on the meta-gradient.  

\subsection{Experimental Results}
\begin{figure}
  \begin{center}
    \includegraphics[width=0.5\textwidth]{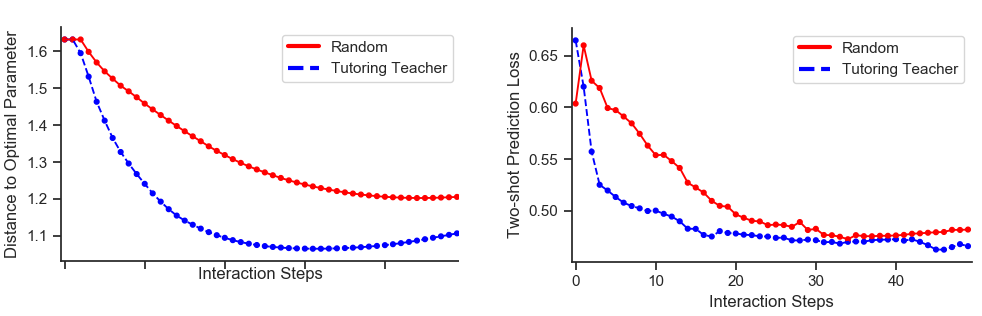}
  \end{center}
  \caption{Top: by optimising the choice and order of tasks we can guide the online meta-learner towards a good initialization. Bottom: machine education leads to faster improvements on two-shot prediction loss for the online meta-learner.}
  \label{fig:meta_learning_sinusoidal}
\end{figure}
\paragraph{Setup.} 
We generated 100 randomly selected non-linear regression tasks by using the class of sine functions as described in \cite{finn2017model}. The meta-learner employs a neural network and we aim to find a good initialization $\theta_0 \in \Theta$ for this network. Here, $\Theta$ is a real-valued vector space and $d(.,.)$ is the Euclidean distance.
We first trained a neural network to perform regression using all 100 tasks with model-agnostic meta-learning and took the resulting initialization of this offline-trained neural network as $\theta^*$, the optimal network initialization we would like to guide a learner towards. 
The learner employs the online meta-learning method with the follow-the-meta-leader algorithm~\cite{finn2019online}. 
We have limited the number of tasks to 50, where the online meta-learner receives 50 tasks from the set of 100 training tasks sequentially. All experiments are conducted with 10 seeds and mean results are reported. Standard deviations are provided in the supplement.
    
\paragraph{Result.} 
Figure \ref{fig:meta_learning_sinusoidal} shows that machine education is able to guide the online meta-learner towards $\theta^*$, which leads to quick improvements in 2-shot prediction loss compared to random task selection. 
The 2-shot prediction loss is evaluated by a test task the network has never seen before, randomly sampled from the distribution over sine functions. 


\section{Related works.} The proposed framework is closely related to the problem of sequential machine teaching. Machine teaching~\cite{zhu2015machine, GOLDMAN199520} addresses the inverse problem of machine learning, where a teacher must select an optimal dataset to present to a learner. A machine teaching method aims to select a minimal dataset $D$ such that the model $\theta = Alg(D)$ learned by a machine learner based on algorithm $Alg$ is close to an optimal model~$\theta^*$ ~\cite{zhu2018overview}. An iterative variant~\cite{liu2017iterative} assesses the iterative nature of some learning algorithms and shifts the problem from minimizing the size of a dataset to minimizing the number of steps. However, this method still assumes that the learner is fully-observed by the teacher (in particular that the learning algorithm is known) and that the teacher can only exchange data points. 
The method introduced by~\citet{liu2017towards} alleviates these two problems, by considering that the learner and the teacher have different views of the same data and that the teacher does not know the algorithm of the learner, in a same way as proposed for the batch-version in~\cite{pmlr-v97-dasgupta19a}. The choice of different views for the learner and the teacher is still different from what we propose, since we do not restrict the actions of the teacher to the choice of data points. More importantly, the main difference is that they consider an unobserved but fixed and unchanged algorithm for the learning, while our setting is built upon the possibility for the teacher to affect the algorithm of the learner.
While~\citet{liu2017towards} apply gradient-based methods, other alternatives have been proposed, based for instance on optimal control~\cite{Lessard19optimalControl}, or models for sequential tasks where the learner is an inverse reinforcement learner~\cite{cakmak2012algorithmic, haug2018teaching, parameswaran2019interactive, tschiatschek2019learner}. A multi-agent formulation has been proposed by \citet{hadfield2016cooperative} for teaching inverse reinforcement learners. 
In all these methods, the learner adapts to the teacher by updating only their estimated model and this line of work considers only the states of the world, whereas in our work we take one step further to considering the teacher's influence on the inner states of the learner (e.g. its priors, learning rate...) which affects \textit{both} the learner's model and their learning algorithm. 
Finally, \citet{peltola2019machine} proposed manipulative teaching of active sequential learners, where a manipulative teacher can steer the learner towards the parameters of its liking 
and showed that manipulation is more effective if the teacher has a model of the learner. However, this teaching strategy cannot achieve generalization on future tasks.

Multiple human teaching tasks have been formulated in terms of MDPs or POMDPs. In particular, the method proposed by~\citet{fan2018learning2teach} considers that the teacher uses an MDP to adapt its teaching policy to the learner during the teaching process.   In the domain of Intelligent Tutoring Systems, the use of multi-arm bandits has been suggested by~\citet{clement2015bandits} as a way to adapt to multiple types of learners. As an alternative, POMDPs have been proposed to alleviate the uncertainty over the learner's cognitive state~\cite{rafferty2016faster}. Unlike our method, these papers only consider adapting to various profiles of learners, but do not consider the possibility of switching from one to another. 

\section{Discussion}

We proposed machine education as a generalization of machine teaching to learners with inner states, which aims at ``enlightening'' learners while teaching them by considering their independent learning performance in the future. Our framework extends from traditional machine teaching to including learners who are learning to learn, human and machine alike. Beyond its applications in pedagogical tools, this setting opens various questions that were not yet addressed in this paper. From a theoretical point of view, it extends the question of teaching dimension to the minimal number of interactions necessary to teach in a non-manipulative way. For practical applications, we introduced a general setting but the question of how to design the learner's types and transition dynamics remains partially open; besides manually tailored solutions for each task, the models could be learned from off-line collected interactions building on simplified task models. 
We provide a further discussion on this important issue in the supplementary material.

\section*{Acknowledgments}

This work was supported by the Academy of Finland (Flagship programme: Finnish Center for Artificial Intelligence, FCAI, and grants 328400, 319264, 292334). Mustafa Mert \c Celikok is partially funded by a personal grant from the Finnish Science Foundation for Technology and Economics (KAUTE). We acknowledge the computational resources provided by the Aalto Science-IT Project.

\bibliography{biblio}

\section*{Broader Impact}

The proposed contribution can be seen from two different perspectives: teaching of machines and teaching of humans. Teaching of machines is intrinsically related to meta-learning and to the possibility of making a machine learner able to choose its algorithm by itself. 

In the context of teaching human learners, which is on the rise with the emergence of Intelligent Tutoring Systems (ITS)~\cite{du2016artificial}, the question of designing high-quality artificial teachers is a priority. However, as exposed in~\cite{cochran2003teaching}, even if there is a consensus on the need for good-quality teachers, the characteristics of good teaching are less clear. In a public opinion poll~\cite{hart2002national}, it has been observed that only 19\% of the participants mentioned that good-quality teaching entailed for the teacher to have a thorough understanding of the subject, against 42\% for designing learning activities that inspired pupil interest. This observation highlights the perceived importance of pedagogy and points out that a teacher with only excellent knowledge would not be sufficient. The proposed framework alleviates this question, based on three considerations: (1) The thorough understanding of the subject is modeled by the access to $\theta^*$, but teaching $\theta^*$ to the learner is not the sole priority unlike in standard machine teaching for instance; (2) The teacher plans a sequence of interactions with the learner, which corresponds to an understanding of teaching in the long-term; (3) The priority of the teacher is to help the learner progressing in their understanding. Even if the framework we propose is preliminary and cannot be directly applied to ITS, it still paves the way for high-quality automatic teaching. 
An important consideration is the conception of the learner's models, which needs to be learned automatically from observed interactions, or designed by human experts. An inaccurate choice for the model family can have harmful consequences, since seemingly innocent advice may lead to unexpected behaviours. As an illustration, the study proposed in~\cite{mcnee2006don} shows that one irrelevant recommendation is enough to lose the trust of the user: Such a phenomenon would be of dramatic importance in a context of teaching. 

Teaching human learners cannot be limited to interactive tutoring systems though. The example developed in Sections~\ref{sec:motivating-example} and \ref{sec:application-mme} illustrates the possibility of advanced modelling tools for scientists who are not expert statisticians but use statistical analysis to draw conclusions from data. Such assistants could help scientists design statistical models by identifying the need of technical explanations and by sorting the relevant information from the data. In these domains, guaranteeing a non-manipulative teaching is of major importance, so that the users can gain and maintain a perfect understanding of their data. As such, the problem is very close to the question of understandability of Automatic ML (AutoML). Recent studies show that interpretability and visualization are key elements requested by users of AutoML systems~\cite{drozdal2020trust}. Our system would increase the understandability of such systems by making the users participate to the choice of the model and providing them explanations on modeling. 

Finally, even if our work takes the direction of a non-manipulative teaching, we are still far from being able to protect learners against manipulative teachers. We notice that, in our framework, manipulation depends on the state and that, consequently, each learner inner state is associated to a non-manipulative model. Even if we can guarantee to detect a naive manipulative teacher who would impose a model by force by selecting data, we have no guarantee over a teacher who would partially educate. This teacher would adapt their target model $\theta^*$ to pretend being non-manipulative.

\end{document}